\pdfoutput=1

\documentclass[11pt]{article}

\usepackage{naacl2021}

\usepackage{times}
\usepackage{latexsym}

\usepackage[T1]{fontenc}

\usepackage[utf8]{inputenc}

\usepackage{microtype}

\usepackage{microtype}
\usepackage{booktabs}
\usepackage{amsmath}
\usepackage{flexisym}
\usepackage{dcolumn}
\usepackage{kotex}
\usepackage{hhline}
\usepackage{verbatim}
\usepackage{multirow}
\usepackage{amssymb}
\usepackage{rotating}
\usepackage{subfigure}
\usepackage{dblfloatfix}
\usepackage{enumitem}
\usepackage{framed}
\usepackage[utf8]{inputenc}
\usepackage[english]{babel}
\usepackage{xcolor}
\usepackage{tabulary}
\usepackage{color}
\usepackage{arydshln}

\newcolumntype{L}[1]{>{\raggedright\let\newline\\\arraybackslash\hspace{0pt}}m{#1}}
\newcolumntype{C}[1]{>{\centering\let\newline\\\arraybackslash\hspace{0pt}}m{#1}}
\newcolumntype{R}[1]{>{\raggedleft\let\newline\\\arraybackslash\hspace{0pt}}m{#1}}

\newcommand{\jb}[1]{\textcolor{blue}{#1}}
\newcommand{\henry}[1]{\textcolor{black}{#1}}

\title{KPQA: A Metric for Generative Question Answering\\ Using Keyphrase Weights}

\author{Hwanhee Lee$^{1}$, Seunghyun Yoon$^{2}$, Franck Dernoncourt$^{2}$ \\
\bf Doo Soon Kim$^{3}$\thanks{\hspace{0.1cm} This research was done while the author was affiliated with Adobe Research.}, Trung Bui$^{2}$, Joongbo Shin$^{1}$~\and Kyomin Jung$^{1}$ \\
$^{1}$Dept. of Electrical and Computer Engineering, Seoul National University, Seoul, Korea \\
$^{2}$Adobe Research, San Jose, CA, USA, $^{3}$Roku Inc., San Jose, CA, USA\\
{\tt \{wanted1007, jbshin, kjung\}@snu.ac.kr} \\
{\tt \{syoon, franck.dernoncourt, bui\}@adobe.com, dskim@roku.com}\\
}

\begin{document}
\maketitle
\begin{abstract}
In the automatic evaluation of generative question answering (GenQA) systems, it is difficult to assess the correctness of generated answers due to the free-form of the answer. Especially, widely used n-gram similarity metrics often fail to discriminate the incorrect answers since they equally consider all of the tokens. To alleviate this problem, we propose KPQA-metric, a new metric for evaluating the correctness of GenQA. Specifically, our new metric assigns different weights to each token via keyphrase prediction, thereby judging whether a generated answer sentence captures the key meaning of the reference answer. To evaluate our metric, we create high-quality human judgments of correctness on two GenQA datasets. 
Using our human-evaluation datasets, we show that our proposed metric has a significantly higher correlation with human judgments than existing metrics. Code for KPQA-metric will be available at \url{https://github.com/hwanheelee1993/KPQA}.
\end{abstract}

\section{Introduction}
Question answering (QA) has received consistent attention from the natural language processing community. Recently, research on QA systems has reached the stage of generating free-form answers, called GenQA, beyond extracting the answer to a given question from the context \cite{yin2016neural, song2017unified, bauer2018commonsense, nishida2019multi, bi2019incorporating, bi2020generating}.
However, as a bottleneck in developing GenQA models, there are no proper automatic metrics to evaluate generated answers~\cite{chen2019evaluating}.
\begin{figure}[!t]
\small

\begin{framed}
\textbf{Context} : ... , this process, called hypothesis testing, consists of \textbf{\textcolor{red}{four steps}}. , ...\\\\
\textbf{Question} : \textbf{\textcolor{red}{How many steps}} are involved in a hypothesis test?\\
\textbf{Reference Answer} : \textbf{\textcolor{red}{Four steps}} are involved in a hypothesis test.\\
\textbf{Generated Answer} : There are \textbf{\textcolor{red}{seven steps}} involved in a hypothesis test .\\\\
\textbf{Human Judgment} : 0.063\\\\
\textbf{BLEU-1} : 0.778 \hspace{10.5mm} \textbf{BLEU-1-KPQA} : 0.057\\
\textbf{ROUGE-L} : 0.713 \hspace{7mm} \textbf{ROUGE-L-KPQA} : 0.127
\end{framed}
\caption{
An example from MS-MARCO~\cite{bajaj2016ms} where widely used n-gram similarity metrics does not align with human judgments of correctness. On the other hand, our \textit{KPQA}-metrics focus on the key information and give low scores to incorrect answers similar to humans.
} 

\label{fig_ex}
\end{figure}

In evaluating a GenQA model, it is essential to consider whether a generated response correctly contains vital information to answer the question.
There exist several n-gram similarity metrics such as BLEU~\cite{papineni-etal-2002-bleu} and ROUGE-L~\cite{lin-2004-rouge}, that measure the word overlaps between the generated response and the reference answer; however, these metrics are insufficient to evaluate a GenQA system \cite{yang2018adaptations, chen2019evaluating}. 

For instance, in the example in Figure~\ref{fig_ex} from the MS-MARCO~\cite{bajaj2016ms}, the generated answer receives a high score on BLEU-1 (0.778) and ROUGE-L (0.713) due to the many overlaps of words with those in the reference. 
However, humans assign a low score of 0.063 on the scale from 0 to 1 due to the mismatch of critical information.
As in this example, we find that existing metrics often fail to capture the correctness of the generated answer that considers the key information for the question.

To overcome this shortcoming of the existing metrics, we propose a new metric called KPQA-metric for evaluating GenQA systems. To derive the metric, we first develop Keyphrase Predictor for Question Answering (KPQA). KPQA computes the importance weight of each word in both the generated answer and the reference answer by considering the question.
By integrating the output from the KPQA, we compute the KPQA-metric in two steps: (1) Given a \{\textit{question}, \textit{generated answer}, \textit{reference answer}\}, we compute importance weights for each question-answer pair \{\textit{question}, \textit{generated answer}\} and \{\textit{question}, \textit{reference answer}\} using a KPQA; (2) We then compute a weighted similarity score by integrating the importance weights into existing metrics. Our approach can be easily integrated into most existing metrics, including n-gram similarity metrics and the recently proposed BERTScore~\cite{zhang2020bertscore}.

Additionally, we newly create two datasets for assessing automatic evaluation metrics with regard to the correctness in the GenQA domain. We first generate answers using state-of-the-art GenQA models on MS-MARCO and AVSD~\cite{alamri2019audio} where the target answers are natural sentences rather than short phrases. We then collect human judgements of correctness over the 1k generated answers for each dataset.

In experiments on the human-evaluation datasets, we show that our KPQA-metrics have significantly higher correlations with human judgments than the previous metrics. For example, BERTScore-KPQA, one of our KPQA-integrated metrics, obtains Pearson correlation coefficients of 0.673 on MS-MARCO whereas the original BERTScore obtains 0.463. Further analyses demonstrate that our KPQA-metrics are robust to the question type and domain shift.
Overall, our main contributions can be summarized as follows: 
\begin{itemize}[noitemsep,leftmargin=*]
\item We propose KPQA metric, an importance weighting based evaluation metric for GenQA.
\item We collect high-quality human judgments of correctness for the model generated answers on MS-MARCO and AVSD, where those two GenQA datasets aim to generate sentence-level answers. We show that our proposed metric has a dramatically higher correlation with human judgments than the previous metrics for these datasets.
\item We verify the robustness of our metric in various aspects such as question type and domain effect. 
\item We release the human-annotated benchmark dataset and pre-trained models to compute the KPQA-metric to the research community\footnote{https://github.com/hwanheelee1993/KPQA}.
\end{itemize}
\section{Preliminaries: Automated Text Evaluation Metrics}
We briefly review the current automated text evaluation metrics that have been used to evaluate GenQA systems.

\noindent \textbf{BLEU} is a popular evaluation metric for generated text based on $n$-gram precision. BLEU scores a candidate by counting the number present in the reference among the $n$-gram of the candidate. In general, $n$ varies from 1 to 4, and the scores for varying $n$ are aggregated with a geometric mean. 

\noindent \textbf{ROUGE} is a set of evaluation metrics used for automatic text generation such as summarization and machine translation. Typically, most studies use ROUGE-L, which is a F-measure based on the longest common subsequence between a candidate and the reference.

\noindent \textbf{METEOR}~\cite{banerjee2005meteor} is an F1 score of a set of unigram alignments. METEOR has a unique property that it considers stemmed words, synonyms, and paraphrases, as well as the standard exact word matches.

\noindent \textbf{CIDER}~\cite{vedantam2015cider} is a consensus-based evaluation metric that is designed for a high correlation with human judgment in the image captioning problem. CIDEr uses Term Frequency-Inverse Document Frequency (TF-IDF) weights for human-like evaluation.

\noindent \textbf{BERTScore} is a recently proposed text evaluation metric that use pre-trained representations from BERT~\cite{devlin2019bert}. BERTScore first computes the contextual embeddings for given references and candidates independently with BERT, and then computes pairwise cosine similarity scores. When computing similarity, BERTScore adopts Inverse Document Frequency (IDF) to apply importance weighting.

\begin{figure*}[t]
\small
\centering
\includegraphics[trim=0 0 0 0, clip, width=2\columnwidth]{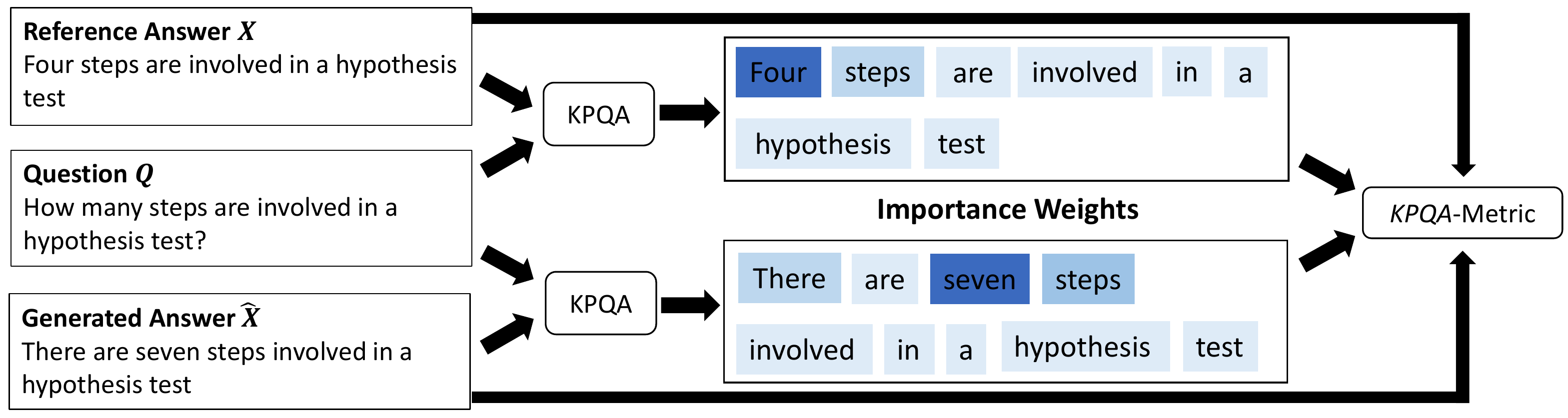}
\caption{
Overall flow of KPQA-metric. Importance weights are computed by pre-trained KPQA for each question-answer pair. And then these weights are integrated into existing metrics to compute weighted similarity.}
\label{fig_overall}
\end{figure*}

\section{Proposed Metric for Evaluating GenQA}
\label{sec:importance}
To build a better metric for GenQA, we first propose KPQA.
By considering the question, the KPQA assigns different weights to each token in the answer sentence such that salient tokens receive a high value.
We then integrate the KPQA into existing metrics to make them evaluate correctness as well.
\begin{figure}[t]
\small
\centering
\includegraphics[width=1.0\columnwidth]{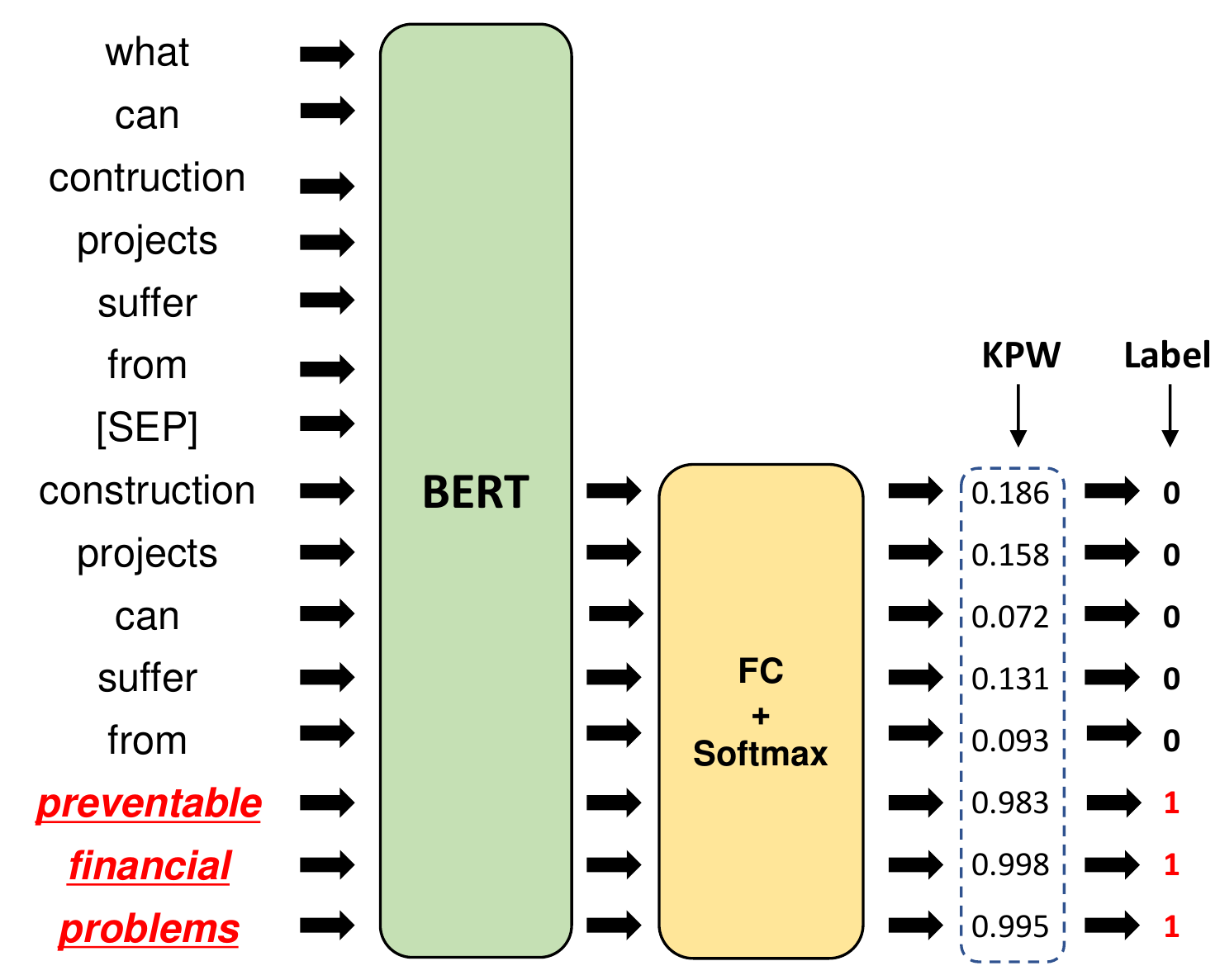}
\caption{
Overall architecture and an output example of KPQA. KPQA classifies whether each word in the answer sentences is in the answer span for a given question. We use the output probability KPW as an importance weight to be integrated into KPQA-metric.
}
\label{fig_kpqa_all}
\end{figure}

\subsection{KPQA}
For GenQA, we observe that each word has different levels of importance when assessing a generated answer.
As shown in Figure~\ref{fig_ex}, there exist keywords or keyphrases that are considered significant when evaluating the correctness of the answer. Additionally, some words, such as function words are mostly irrelevant to the correctness of the answer. Inspired by this observation, we introduce KPQA, which can predict the importance of each word when evaluating GenQA systems.
As shown in Figure~\ref{fig_kpqa_all}, KPQA is a BERT-based~\cite{devlin2019bert} classifier that predicts salient tokens in the answer sentences depending on the question.
We regard it as a multi-class classification task where each token is a single class.
To train KPQA, we first prepare extractive QA datasets such as SQuAD~\cite{rajpurkar2016squad}, which consist of \{\textit{passage}, \textit{question}, \textit{answer-span}\}.
We transform these datasets into pairs of \{\textit{answer-sentences}, \textit{question}, \textit{answer-span}\}.
We extract the answer-sentences that contain answer-span in the passage since these sentences are short summaries for the given question. Specifically, for a single-hop QA dataset such as SQuAD, we pick a single sentence that includes answer-span as the answer sentence. For the answers in a multi-hop QA dataset such as HotpotQA~\cite{yang2018hotpotqa}, there are multiple supporting sentences for the single answer span. For these cases, we use SpanBERT~\cite{joshi2020spanbert} to resolve the coreferences in the paragraphs and extract all of the supporting sentences to compose answer sentences.
The \{\textit{question}, [SEP], \textit{answer-sentences}\} is then fed into the KPQA to classify the answer-span, which is a set of salient tokens, in the given answer-sentences considering the question.

\subsection{KPQA Metric}
Since KPQA's training process allows KPQA to find essential words in the answer sentences to a given question, we use a pre-trained KPQA to get the importance weights that are useful for evaluating the correctness of generated answers in GenQA.
The overall flow of our KPQA-metric is described in Figure~\ref{fig_overall}.
We describe how we combine these weights with existing metrics to derive the KPQA-metric.

We first compute the importance weights for a given question ${Q}$ = ($q_1$, ..., $q_l$), reference answer ${X}$ = (${x}_1$, ..., ${x}_n$) and generated answer $\hat{X}$ =  ($\hat{x}_1$, ..., $\hat{x}_m$) using pre-trained KPQA.
We provide each pair \{\textit{question}, \textit{generated answer}\} and \{\textit{question}, \textit{reference answer}\} to pre-trained KPQA and get the output of the softmax layer. 
We define these parts as KeyPhrase Weight (KPW) as shown in Figure~\ref{fig_kpqa_all}. We note that $\text{KPW}^{(Q, \mathbf{\textit{\^{X}}} )}$ = ($w_1$, ..., $w_m$) is an importance weight of generated answer $\hat{X}$ for a given question ${Q}$. These weights reflect the importance of each token for evaluating the correctness.

We then compute KPQA-metric by incorporating the KPW into several existing metrics modifying the precision and recall to compute the weighted similarity.  
 
\paragraph{BLEU-1-KPQA:} We derive BLEU-1-KPQA, which is an weighted precision of unigram ($P_{Unigram}^{KPQA}$) as follows:

\begin{equation}
\begin{aligned}
    P_{Unigram}^{KPQA} = \frac{\Sigma_{i=1}^{m}\Sigma_{j=1}^{n}   \text{KPW}_i^{(Q, \mathbf{\textit{\^{X}}} )} \cdot I(i,j)}
{\Sigma_{i=1}^{m} \text{KPW}_i^{(Q, \mathbf{\textit{\^{X}}} )}},\\
\end{aligned}
\end{equation}
where $I(i,j)$ is an indicator function assigned the value of 1 if token $x_i$ is the same as $\hat{x}_j$ and 0 otherwise.

\paragraph{ROUGE-L-KPQA:} We also derive ROUGE-L-KPQA, which is a modified version of ROUGE-L using KPW to compute weighted precision($P_{LCS}^{KPQA}$), recall($R_{LCS}^{KPQA}$) and F1($F1_{LCS}^{KPQA}$), as follows:

\begin{equation}
\begin{aligned}
    P_{LCS}^{KPQA} = \frac{LCS^{KPQA}(\mathbf{\textit{X}}, \mathbf{\textit{\^{X}}}) }
{\Sigma_{i=1}^{m} \text{KPW}_i^{(Q, \mathbf{\textit{\^{X}}} )}},
\end{aligned}
\end{equation}
\begin{equation}
\begin{aligned}
    R_{LCS}^{KPQA} = \frac{LCS^{KPQA}(\mathbf{\textit{X}}, \mathbf{\textit{\^{X}}}) }
{\Sigma_{i=1}^{n} \text{KPW}_i^{(Q, \mathbf{\textit{X}})}},
\end{aligned}
\end{equation}
\begin{equation}
\begin{aligned}
    F_{LCS}^{KPQA} = \frac{(1 +\beta^2)R_{LCS}^{KPQA} P_{LCS}^{KPQA} }
{R_{LCS}^{KPQA} + \beta^2 P_{LCS}^{KPQA}},\\
\end{aligned}
\end{equation}
where LCS is the Longest Common Subsequence between a generated answer and a reference answer.
The $LCS^{KPQA}(\mathbf{\textit{X}}, \mathbf{\textit{\^{X}}})$ is defined as follows:
\begin{equation}
    LCS^{KPQA}(\mathbf{\textit{X}}, \mathbf{\textit{\^{X}}})= \Sigma_{i=1}^{m} I_i \cdot \text{KPW}_i^{(Q,\mathbf{\textit{ \^{X}}})},\\
\end{equation}
where $I_i$ is an indicator function which is 1 if each word is in the LCS and 0 otherwise. $\beta$ is defined in~\cite{lin-2004-rouge}.

\paragraph{BERTScore-KPQA} Similar to ROUGE-L-KPQA, we compute BERTScore-KPQA using KPW. We first compute contextual embedding $\mathbf{\hat{x}}$ for generated answer $\hat{X}$ and $\mathbf{x}$ for reference ${X}$ using the BERT model. Then, we compute weighted precision($P_{BERT}^{KPQA}$), recall($R_{BERT}^{KPQA}$) and F1($F1_{BERT}^{KPQA}$) with contextual embedding and KPW of each token as follows:

\begin{equation}
\begin{aligned}
    P_{BERT}^{KPQA} = \frac{\Sigma_{i=1}^{m}  \text{KPW}_i^{(Q, \mathbf{\textit{\^X}} )} \cdot \text{max}_{x_j\in x} \mathbf{x_i}^T\mathbf{\hat{x}_j} }
{\Sigma_{i=1}^{m} \text{KPW}_i^{(Q, \mathbf{\textit{ \^{X}}})}}\\
\end{aligned}
\end{equation}

\begin{equation}
\begin{aligned}    
    R_{BERT}^{KPQA} = \frac{\Sigma_{i=1}^{n} \text{KPW}_i^{(Q, \mathbf{\textit{X}} )} \cdot \text{max}_{\hat{x}_j\in \hat{x}} \mathbf{x_i}^T\mathbf{\hat{x}_j} }
{\Sigma_{i=1}^{n} \text{KPW}_i^{(Q, \mathbf{\textit{X}})}}\\
\end{aligned}
\end{equation}

\begin{equation}
\begin{aligned}    
    F1_{BERT}^{KPQA} = 2\cdot \frac{P_{BERT}^{KPQA}\cdot R_{BERT}^{KPQA}}{P_{BERT}^{KPQA}+R_{BERT}^{KPQA}}
\end{aligned}
\end{equation}

\begin{equation}
\begin{aligned}
    P_{LCS}^{KPQA} = \frac{LCS^{KPQA}(\mathbf{\textit{X}}, \mathbf{\textit{\^{X}}}) }
{\Sigma_{i=1}^{m} \text{KPW}_i^{(Q, \mathbf{\textit{\^{X}}} )}},
\end{aligned}
\end{equation}
\begin{equation}
\begin{aligned}
    R_{LCS}^{KPQA} = \frac{LCS^{KPQA}(\mathbf{\textit{X}}, \mathbf{\textit{\^{X}}}) }
{\Sigma_{i=1}^{n} \text{KPW}_i^{(Q, \mathbf{\textit{X}})}},
\end{aligned}
\end{equation}
\begin{equation}
\begin{aligned}
    F_{LCS}^{KPQA} = \frac{(1 +\beta^2)R_{LCS}^{KPQA} P_{LCS}^{KPQA} }
{R_{LCS}^{KPQA} + \beta^2 P_{LCS}^{KPQA}},\\
\end{aligned}
\end{equation}
where LCS is the Longest Common Subsequence between a generated answer and a reference answer.
The $LCS^{KPQA}(\mathbf{\textit{X}}, \mathbf{\textit{\^{X}}})$ is defined as follows:
\begin{equation}
    LCS^{KPQA}(\mathbf{\textit{X}}, \mathbf{\textit{\^{X}}})= \Sigma_{i=1}^{m} I_i \cdot \text{KPW}_i^{(Q,\mathbf{\textit{ \^{X}}})},\\
\end{equation}
where $I_i$ is an indicator function which is 1 if each word is in the LCS and 0 otherwise. $\beta$ is defined in~\cite{lin-2004-rouge}. \henry{Similar to ROUGE-L-KPQA, we also derive BLEU-1-KPQA and BERTScore-KPQA by intergating KPW and provide the formulas in Appendix.}

\section{Collecting Human Judgments}

\subsection{Generating Answers}
\label{sec:dataset}

\begin{table}[h]
\small
\centering
\begin{tabular}
{C{0.4\columnwidth}C{0.25\columnwidth}C{0.18\columnwidth}}
\toprule
\textbf{\ Dataset} & \textbf{Answer Length (avg.)} & \textbf{\# Samples}\\
\midrule
MS MARCO & 16.6 & 183k  \\
AVSD & 9.4 & 118k \\
\hdashline
Narrative QA & 4.7 & 47k \\
SemEval & 2.5 & 14k \\ 
\bottomrule 
\end{tabular}
\caption{
Statistics of the generative question answering dataset.
}
\label{anslen}
\end{table}
\paragraph{GenQA Datasets:} 
To evaluate GenQA metrics, it is necessary to measure the correlation between human judgments and automated text evaluation metrics for evaluating the model generated answers.
Recently, \citeauthor{chen2019evaluating} (2019) released human judgments of correctness for two GenQA datasets, NarrativeQA~\cite{kovcisky2018narrativeqa} and SemEval-2018 Task 11 (SemEval)~\cite{ostermann2018semeval}. 
However, we find that the average lengths of the answer sentence are 4.7 and 2.5 for NarrativeQA and SemEval, respectively, as shown in Table~\ref{anslen}. 
These short answers are often short phrases and cannot be representative of GenQA, because the answers could be long and may deliver complex meaning.
We argue that evaluating long and abstractive answers is more challenging and suitable for studying the metrics for general form of GenQA.
To fill this gap, we collect the human judgments of correctness for model generated answers on two other GenQA datasets, MS-MARCO and AVSD, which have longer answers than NarrativeQA and SemEval as shown in Table~\ref{anslen}. For the MS-MARCO, we use the Natural Language Generation (NLG) subset, which has more abstractive and longer answers than the Q\&A subset.

\paragraph{GenQA Models:} 
For each of the two datasets, we first generate answers for questions on validation sets using two trained GenQA models: UniLM~\cite{dong2019unified} and MHPGM~\cite{bauer2018commonsense} for MS-MARCO, MTN~\cite{le2019multimodal} and AMF~\cite{alamri2018audio, hori2017attention} for AVSD.
Details on these QA models are in Appendix. 
After training, we select 1k samples for each dataset in the validation set. Specifically, we first randomly pick the 500 questions in the validation set of each dataset and collect the corresponding model generated answers for each model so that we have two generated answers for each sample. Therefore, we collect a total of 1k samples, two different answers for 500 questions for each dataset. Also, we discard samples if one of two GenQA models exactly generates the ground-truth answer since human evaluation is useless during the sampling. 

\subsection{Collecting Human Judgments of Answer Correctness}
\label{sec:collect}
We hire workers from the Amazon Mechanical Turk (MTurk) to rate the correctness of the generated answers from the models we trained. We assign ten workers for each sample to get reliable data. We ask the workers to annotate correctness using a 5-point Likert scale~\cite{likert1932technique}, where 1 means completely wrong, and 5 means completely correct. We provide the full instruction in Appendix.

\paragraph{Filtering Noisy Workers:} 
Some workers did not follow the instructions, producing poor-quality judgments.
To solve this problem, we filter noisy ratings using the z-score, as in~\cite{jung2011improving}. We first compute the z-score among the ten responses for each sample. Then, we consider the responses whose z-score is higher than 1 to be noise and remove up to five of them in the order of the z-score. 
The average number of annotators after filtering is shown in Table~\ref{inter_agree}. We use the average score of the annotators for each sample as a ground-truth evaluation score to assess the quality of the evaluation metric.

\begin{table}[t]
\small
\centering
\begin{tabular}
{C{0.3\columnwidth}C{0.2\columnwidth}C{0.33\columnwidth}}
\toprule
\textbf{Dataset} & \textbf{$\alpha$} & \textbf{\# Annotators (avg.)} \\
\midrule
MS MARCO & 0.817 & 7.08 \\ 
AVSD & 0.725 & 6.88 \\ 
\bottomrule 
\end{tabular}
\caption{
Inter annotator agreement measured by Krippendorff's alpha($\alpha$) and the average of number of annotators for each dataset. 
}
\label{inter_agree}
\end{table}
\paragraph{Inter-Annotator Agreement:}
The final dataset is further validated with Krippendorff's alpha~\cite{krippendorff1970estimating, krippendorff2011computing}, a statistical measure of inter-rater agreement for multiple annotators.
We observe that Krippendorff's $\alpha$ is higher than 0.6 for both datasets and models after filtering, as shown in Table~\ref{inter_agree}. These coefficient numbers indicate a ``substantial`` agreement according to one of the general guidelines~\cite{landis1977measurement} for kappa-like measures.

\begin{table*}[!t]
\small
\centering
\begin{tabular}
{L{0.37\columnwidth}C{0.15\columnwidth}C{0.15\columnwidth}C{0.15\columnwidth}C{0.15\columnwidth}
C{0.15\columnwidth}C{0.15\columnwidth}C{0.15\columnwidth}C{0.15\columnwidth}}
\toprule

\multicolumn{1}{c}{\textbf{Dataset}} & \multicolumn{2}{c}{\textbf{MS-MARCO}} & 
\multicolumn{2}{c}{\textbf{AVSD}} & \multicolumn{2}{c}{\textbf{NarrativeQA}} &
\multicolumn{2}{c}{\textbf{SemEval}}\\ 
\cmidrule{1-9}
 \textbf{Metric}
              & \textbf{$r$} & \textbf{$\rho$} 
              & \textbf{$r$} & \textbf{$\rho$}
              & \textbf{$r$} & \textbf{$\rho$}
              & \textbf{$r$} & \textbf{$\rho$}\\
\midrule
\textbf{BLEU-1} & 0.349 & 0.329 & 0.580 & 0.562 & 0.634 & 0.643 & 0.359 & 0.452 \\
\textbf{BLEU-4} & 0.193 & 0.244 & 0.499 & 0.532 & 0.258 & 0.570 & -0.035 & 0.439 \\
\textbf{ROUGE-L} & 0.309 & 0.301 & 0.585 & 0.566 & 0.707 & 0.708 & 0.566 & 0.580 \\
\textbf{METEOR} & 0.423 & 0.413 & 0.578 & 0.617 & 0.735 & 0.755 & 0.543 & 0.645 \\
\textbf{CIDEr} & 0.275 & 0.278 & 0.567 & 0.600 & 0.648 & 0.710 & 0.429 & 0.595 \\
\textbf{BERTScore} & 0.463 & 0.456 & 0.658 & 0.650 & \textbf{0.785} & 0.767 & 0.630 & 0.602 \\
\midrule
\textbf{BLEU-1-KPQA} & 0.675 & 0.634 & 0.719 & 0.695 & 0.716 & 0.699 & 0.362 & 0.462 \\
\textbf{ROUGE-L-KPQA} & \textbf{0.698} & 0.642 & 0.712 & 0.702 & 0.774 & 0.750 & \textbf{0.742} & \textbf{0.687} \\
\textbf{BERTScore-KPQA} & 0.673 & \textbf{0.655} & \textbf{0.729} & \textbf{0.712} & 0.782 & \textbf{0.770} & 0.741 & 0.676 \\


\bottomrule 
\end{tabular}
\caption{
\henry{Pearson Correlation($r$) and Spearman's Correlation($\rho$) between various automatic metrics and human judgments of correctness. All of the results are statistically significant (p-value $<$ 0.01).}
}
\label{comparison-all}
\end{table*}

\section{Experiments}
\label{sec:experiments}
\subsection{Implementation Details}
We choose three datasets SQuAD v1.1~\cite{rajpurkar2016squad}, HotpotQA~\cite{yang2018hotpotqa} and MS-MARCO Q\&A subset to train KPQA. We combine the training set of the three datasets and use a 9:1 split to construct the training and development set of KPQA. For HotpotQA, we exclude \textit{yes/no} type questions where the answers are not in the passage.
\newline
\indent For model parameters, we choose \textit{bert-base-uncased} variants for the BERT model and use one fully-connected layer with softmax layer after it. We train 5 epochs and choose the model that shows the minimum evaluation loss. We provide more details in Appendix.

\subsection{Results}

\paragraph{Evaluation Methods for Metrics:}
To compare the performance of various existing metrics and our metric, we use the Pearson coefficient and Spearman coefficient. We compute these correlation coefficients with human judgments of correctness. We test using MS-MARCO, AVSD, from which we collected human judgments, and NarrativeQA and SemEval from \cite{chen2019evaluating}.

\paragraph{Performance Comparison:}
We present the correlation scores for the baseline metrics and KPQA-augmented ones for multiple datasets in Table~\ref{comparison-all}. 
The correlations between human judgment and most of the existing metrics such as BLEU or ROUGE-L are very low, and this shows that those widely used metrics are not adequate to GenQA. Moreover, the performance of existing metrics is especially low for the MS-MARCO, which has longer and more abstractive answers than the other three datasets.

We observe a significantly higher correlation score for our proposed KPQA-metric compared to existing metrics especially for MS-MARCO and AVSD where the answers are full-sentences rather than short phrases. For the NarrativeQA, where existing metrics also have higher correlations, the gap in performance between KPQA-metric and existing metrics is low. We explain this is because the answers in NarrativeQA are often a single word or short phrases that are already keyphrases.

\paragraph{Comparison with IDF:}
The next best metric after our proposed metric is the original BERTScore, which uses contextual embeddings and adopts IDF based importance weighting.
Since IDF is dependent on the word-frequency among the documents, it can assign a lower weight to some important words to evaluate correctness if they frequently occur in the corpus as shown in Table~\ref{fig_qualiative}.
On the other hand, our KPQA integrated metric assigns weights to words in the answer sentence using the context of the question. This approach provides dynamic weights for each word that leads to a better correlation with human evaluation as shown in Table~\ref{comparison-all}.

\begin{table}[t]
\footnotesize
\centering
\begin{tabular}
{L{0.5\columnwidth}C{0.12\columnwidth}C{0.12\columnwidth}}

\toprule

\multicolumn{1}{c}{\textbf{Dataset}} & \multicolumn{2}{c}{\textbf{MS-MARCO}} \\ 
\cmidrule{1-3}
           \textbf{Metric} & \textbf{$r$} & \textbf{$\rho$}\\
\midrule
\textbf{BLEU-1-KPQA} & 0.675 & 0.634 \\
\textbf{ROUGE-L-KPQA} & \textbf{0.698} & 0.642 \\
\textbf{BERTScore-KPQA} & 0.673 & \textbf{0.655} \\
\midrule
\textbf{BLEU-1-KPQA$_{\text{/MARCO}}$} & 0.573 & 0.529 \\
\textbf{ROUGE-L-KPQA$_{\text{/MARCO}}$} & 0.598 & 0.564 \\
\textbf{BERTScore-KPQA$_{\text{/MARCO}}$} & 0.602 & 0.595 \\
\midrule
\textbf{BLEU-1-KP} & 0.629 & 0.589 \\
\textbf{ROUGE-L-KP} & 0.671 & 0.640 \\
\textbf{BERTScore-KP} & 0.657 & 0.649 \\
\bottomrule 
\end{tabular}
\caption{
Ablation studies for our proposed metrics on domain effect and using the question context.
}
\label{ablation}
\end{table}

\subsection{Ablation Study}

\paragraph{Domain Effect:}
\henry{Our KPQA metric computes importance weights using a supervised model}; thus our proposed method may suffer from a domain shift problem. Although our metric is evaluated on out-of-domain datasets except MS-MARCO, we further examine the effect of the domain difference by changing the trainset of KPQA. 
Since we train KPQA with the combination of SQuAD, HotpotQA and MS-MARCO Q\&A, the original KPQA works as in-domain for MS-MARCO. To measure the negative domain effect, we exclude the MS-MARCO Q\&A in the training set of KPQA and measure the performance of KPQA-metric on MS-MARCO. We annotate it ``-KPQA$_{\text{/MARCO}}$" and report the results in Table~\ref{ablation}. This drop shows the effect of the negative domain shift for our KPQA-metric. However, ``-KPQA$_{\text{/MARCO}}$" is still much higher than all of the previous metrics.

\begin{table*}[t]
\centering
\small
\begin{tabular}{rl}
\toprule
\textbf{Context} & ... , it can take 5-20 hours of walking to lose 1 pound ... , ...\\
\textbf{Question} & How long do i need to walk in order to loose a pound ? \\
\midrule
\textbf{Reference} & 
{\footnotesize
\colorbox[rgb]{1, 1, 1}{\strut Walk}
\colorbox[rgb]{1, 1, 1}{\strut for} 
\colorbox[rgb]{1, 1, 1}{\strut 5}
\colorbox[rgb]{1, 1, 1}{\strut to}
\colorbox[rgb]{1, 1, 1}{\strut 20}
\colorbox[rgb]{1, 1, 1}{\strut hours}
\colorbox[rgb]{1, 1, 1}{\strut to}
\colorbox[rgb]{1, 1, 1}{\strut lose}
\colorbox[rgb]{1, 1, 1}{\strut 1}
\colorbox[rgb]{1, 1, 1}{\strut pound}
\colorbox[rgb]{1, 1, 1}{\strut .}
}\\

\textit{IDF} &
{\footnotesize
\colorbox[rgb]{0.407504, 0.533890, 0.893678}{\strut Walk}
\colorbox[rgb]{0.666270, 0.736312, 0.939588}{\strut for}
\colorbox[rgb]{0.538446, 0.636320, 0.916910}{\strut 5}
\colorbox[rgb]{0.709223, 0.769912, 0.947209}{\strut to}
\colorbox[rgb]{0.501681, 0.607561, 0.910387}{\strut 20}
\colorbox[rgb]{0.573355, 0.663629, 0.923103}{\strut hours}
\colorbox[rgb]{0.709223, 0.769912, 0.947209}{\strut to}
\colorbox[rgb]{0.430007, 0.551494, 0.897670}{\strut lose}
\colorbox[rgb]{0.543188, 0.640030, 0.917751}{\strut 1}
\colorbox[rgb]{0.407504, 0.533890, 0.893678}{\strut pound}
\colorbox[rgb]{0.807216, 0.846568, 0.964595}{\strut .}
}\\
\textit{KPW} &
{\footnotesize
\colorbox[rgb]{0.804557, 0.844488, 0.964123}{\strut Walk}
\colorbox[rgb]{0.790843, 0.833760, 0.961690}{\strut for} 
\colorbox[rgb]{0.327355, 0.471193, 0.879458}{\strut 5}
\colorbox[rgb]{0.329924, 0.473202, 0.879914}{\strut to}
\colorbox[rgb]{0.322427, 0.467338, 0.878584}{\strut 20}
\colorbox[rgb]{0.322440, 0.467348, 0.878586}{\strut hours}
\colorbox[rgb]{0.792543, 0.835090, 0.961991}{\strut to}
\colorbox[rgb]{0.794798, 0.836854, 0.962391}{\strut lose}
\colorbox[rgb]{0.795127, 0.837111, 0.962450}{\strut 1}
\colorbox[rgb]{0.797621, 0.839062, 0.962892}{\strut pound}
\colorbox[rgb]{0.807765, 0.846997, 0.964692}{\strut .}
}\\
\midrule
\multicolumn{2}{c}{\textbf{Human Judgment}: 0.94,\quad\textbf{BERTScore}: 0.72,\quad\textbf{BERTScore-KPQA}: 0.93}\\
\textbf{UniLM} & 
{\footnotesize
\colorbox[rgb]{1, 1, 1}{\strut You}
\colorbox[rgb]{1, 1, 1}{\strut need}
\colorbox[rgb]{1, 1, 1}{\strut to}
\colorbox[rgb]{1, 1, 1}{\strut walk}
\colorbox[rgb]{1, 1, 1}{\strut for}
\colorbox[rgb]{1, 1, 1}{\strut 5}
\colorbox[rgb]{1, 1, 1}{\strut to}
\colorbox[rgb]{1, 1, 1}{\strut 20}
\colorbox[rgb]{1, 1, 1}{\strut hours}
\colorbox[rgb]{1, 1, 1}{\strut in}
\colorbox[rgb]{1, 1, 1}{\strut order}
\colorbox[rgb]{1, 1, 1}{\strut to}
\colorbox[rgb]{1, 1, 1}{\strut loose}
\colorbox[rgb]{1, 1, 1}{\strut a}
\colorbox[rgb]{1, 1, 1}{\strut pound}
\colorbox[rgb]{1, 1, 1}{\strut .}
}\\
\textit{IDF} &
{\footnotesize
\colorbox[rgb]{0.606107, 0.689249, 0.928914}{\strut You}
\colorbox[rgb]{0.473781, 0.585736, 0.905437}{\strut need}
\colorbox[rgb]{0.709223, 0.769912, 0.947209}{\strut to}
\colorbox[rgb]{0.407504, 0.533890, 0.893678}{\strut walk}
\colorbox[rgb]{0.666270, 0.736312, 0.939588}{\strut for}
\colorbox[rgb]{0.538446, 0.636320, 0.916910}{\strut 5}
\colorbox[rgb]{0.709223, 0.769912, 0.947209}{\strut to}
\colorbox[rgb]{0.501681, 0.607561, 0.910387}{\strut 20}
\colorbox[rgb]{0.573355, 0.663629, 0.923103}{\strut hours}
\colorbox[rgb]{0.717078, 0.776057, 0.948602}{\strut in}
\colorbox[rgb]{0.321569, 0.466667, 0.878431}{\strut order}
\colorbox[rgb]{0.709223, 0.769912, 0.947209}{\strut to}
\colorbox[rgb]{0.375788, 0.509080, 0.888051}{\strut loose}
\colorbox[rgb]{0.723300, 0.780925, 0.949706}{\strut a}
\colorbox[rgb]{0.407504, 0.533890, 0.893678}{\strut pound}
\colorbox[rgb]{0.807216, 0.846568, 0.964595}{\strut .}
}\\
\textit{KPW} &
{\footnotesize
\colorbox[rgb]{0.807313, 0.846645, 0.964612}{\strut You}
\colorbox[rgb]{0.807115, 0.846489, 0.964577}{\strut need}
\colorbox[rgb]{0.807170, 0.846532, 0.964586}{\strut to}
\colorbox[rgb]{0.806558, 0.846054, 0.964478}{\strut walk}
\colorbox[rgb]{0.793336, 0.835710, 0.962132}{\strut for}
\colorbox[rgb]{0.324331, 0.468827, 0.878921}{\strut 5}
\colorbox[rgb]{0.327016, 0.470928, 0.879398}{\strut to}
\colorbox[rgb]{0.321985, 0.466992, 0.878505}{\strut 20}
\colorbox[rgb]{0.322142, 0.467115, 0.878533}{\strut hours}
\colorbox[rgb]{0.805706, 0.845387, 0.964327}{\strut in}
\colorbox[rgb]{0.806675, 0.846145, 0.964499}{\strut order}
\colorbox[rgb]{0.806553, 0.846050, 0.964477}{\strut to}
\colorbox[rgb]{0.806046, 0.845653, 0.964387}{\strut loose}
\colorbox[rgb]{0.805165, 0.844964, 0.964231}{\strut a}
\colorbox[rgb]{0.805295, 0.845065, 0.964254}{\strut pound}
\colorbox[rgb]{0.807781, 0.847011, 0.964695}{\strut .}
}\\
\bottomrule
\end{tabular}
\caption{
An example of the scores given by humans, BERTScore and BERTScore-KPQA for the samples from MS-MARCO dataset. BERTScore uses IDF and BERTScore-KPQA uses KPW as importance weights to compute score. Heat map shows IDF and KPW, which are normalized between 0 and 1.
}
\label{fig_qualiative}
\end{table*}
\paragraph{Using the Question Context:} 
Our KPQA uses the question as an additional context to predict the keyphrases in the sentence, as shown in Figure~\ref{fig_kpqa_all}. 
To examine the power of utilizing the question information for the keyphrase predictor, we remove the question part from the dataset and train the keyphrase prediction model.
With the newly trained model, we compute the importance weights for words in the target sentence and apply them to BLEU-1, ROUGE-L, and BERTScore.
We call this metric as ``-KP"  and report the results in Table~\ref{ablation}. We observe that ``-KPQA" metric is better than ``-KP" metric for all of the three variants. These results show that training keyphrase predictor to find the short answer candidate in the sentence is effective for capturing the key information in the generated answer, but it is more effective when the question information is integrated.
\begin{figure}[t]
\small
\centering
\includegraphics[width=1.0\columnwidth]{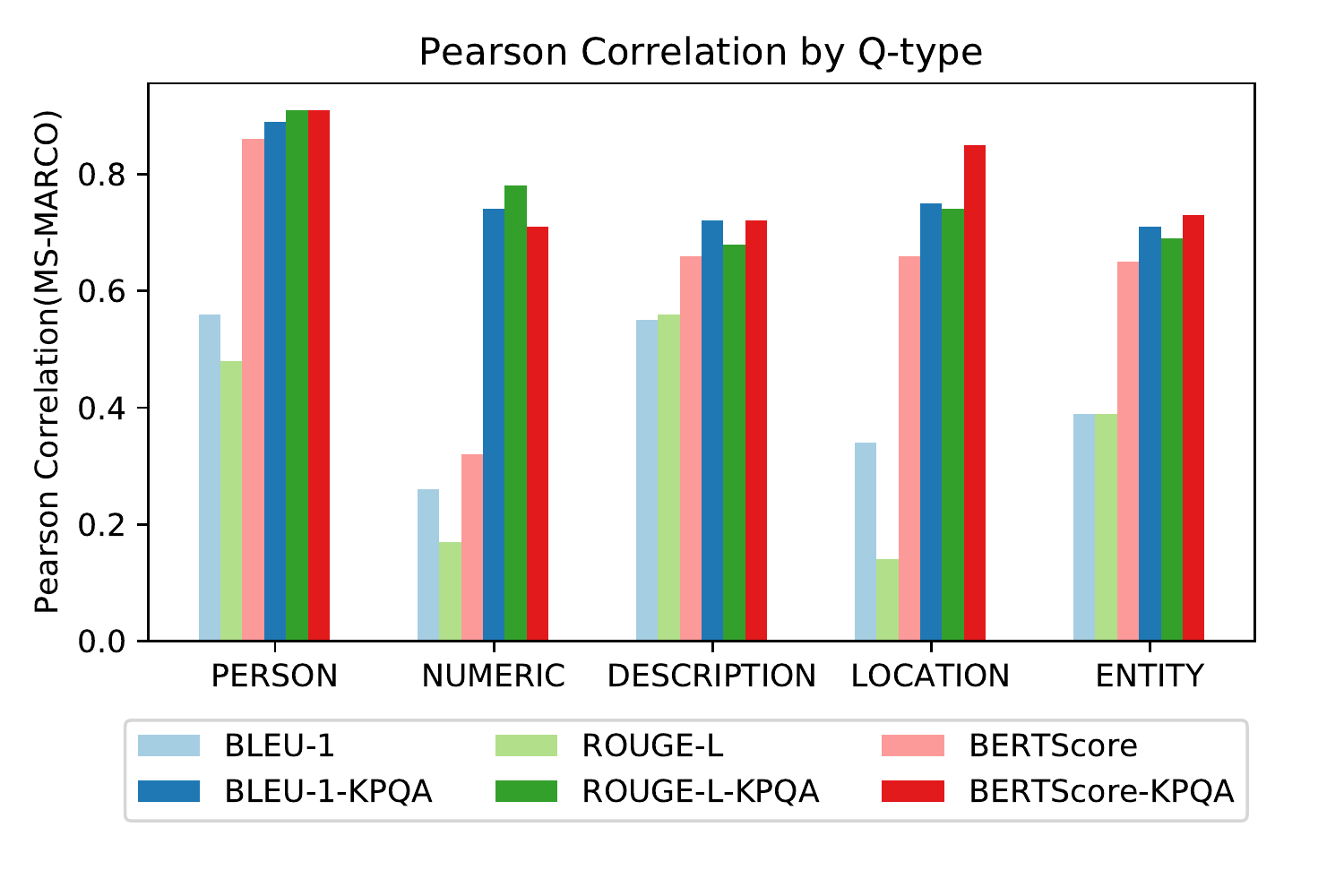}
\caption{
Pearson correlation coefficient among question types on MS-MARCO dataset.
}
\label{fig_cor_qtype}
\end{figure}
\subsection{Analysis}

\paragraph{Correlation Among Question Type:}
\label{paragraph:cor_by_qtype}
Since MS-MARCO provides the question type information  (\textit{PERSON}, \textit{NUMERIC}, \textit{DESCRIPTION}, \textit{LOCATION}, \textit{ENTITY}) for each \{\textit{question}, \textit{answer}\} pair, we evaluate the various metrics by the question type. We split the dataset into these five question types and measure the performance of various metrics with Pearson correlation coefficients. As shown in Figure~\ref{fig_cor_qtype}, our KPQA-metric variants outperform their original version in all of the question types. KPQA-metric is especially effective for the \textit{NUMERIC} question type, whose answer sentence often has shorter keyphrase such as a number. For \textit{ENTITY} and \textit{PERSON} question types, the gap between KPQA-integrated metric and original metric is lower for BERTScore. We speculate that this is because the original BERTScore uses IDF-based importance weighting, unlike other metrics.

\paragraph{Multiple Sentence Answers:}
\begin{figure}[!t]
\small

\begin{framed}
\textbf{Question} : How to cook sausage peppers onions ?\\\\
\textbf{Reference Answer} : To cook sausage peppers onions first place the sausage in a large skillet over medium heat, and brown on all sides after that remove from skillet, and slice meelt butter in the skillet, stir in the yellow onion, red onion, and garlic, and cook 2 to 3 minutes and then mix in red bell pepper and green bell pepper season with basil, and oregano in last stir in white wine.\\\\
\textbf{Generated Answer} : To cook sausage peppers onions , preheat the oven to 350 degrees fahrenheit . Place the onions in the oven and cook for 20 minutes 
\end{framed}

\caption{
An example from MS-MARCO where the answers are composed of multiple sentences.
} 

\label{fig_multi}
\end{figure}

Most of the answers in MS-MARCO and AVSD consist of single sentences, but the answers for GenQA can be multiple sentences like~\cite{fan2019eli5}. To verify our KPQA-metric on multiple sentence answers, we collect additional 100 human judgments for the generated answer whose answers are multiple sentences in the MS-MARCO like the example in Figure~\ref{fig_multi}, and evaluate the various metrics on this dataset. As shown in Table~\ref{table_multi}, our KPQA integrated metric shows still higher correlations than other metrics. We observe that the gap between KPQA integrated metrics and existing metrics is relatively lower than that of Table~\ref{comparison-all}. We speculate this is because many of the multiple sentence answers are \textit{DESCRIPTION} type answers whose keyphrases are sometimes vague, similar to the results in Figure~\ref{fig_cor_qtype}.

\begin{table}[t]
\footnotesize
\centering
\begin{tabular}
{L{0.45\columnwidth}C{0.12\columnwidth}C{0.12\columnwidth}}

\toprule

\multicolumn{1}{c}{\textbf{Dataset}} & \multicolumn{2}{c}{\textbf{MS-MARCO}} \\ 
\midrule
           \textbf{Metric} & \textbf{$r$} & \textbf{$\rho$}\\
\midrule
\textbf{BLEU-1} & 0.363 & 0.364 \\
\textbf{ROUGE-L} & 0.584 & 0.607 \\
\textbf{BERTScore} & 0.712 & 0.728\\
\midrule
\textbf{BLEU-1-KPQA} & 0.529 & 0.540 \\
\textbf{ROUGE-L-KPQA} & 0.642 & 0.648 \\
\textbf{BERTScore-KPQA} & \textbf{0.774} & \textbf{0.786}\\

\bottomrule 
\end{tabular}
\caption{
Correlation coefficients between various automatic metrics and human judgments of correctness for evaluating multiple sentence answers in MS-MARCO~\cite{bajaj2016ms}. 
}
\label{table_multi}
\end{table}

\paragraph{Error Analysis:}
We pick 100 error cases from MS-MARCO in the order of a large difference in ranks among 1k samples between human judgments and BERTScore-KPQA. 
The importance weights have no ground-truth data; thus we manually visualize the weights as shown in Table~\ref{fig_qualiative} and analyze the error cases.

From the analysis, we observe some obvious reasons for the different judgments between humans and BERTScore-KPQA.
We first classify error cases by the question types and observe that 51 cases belong to \textit{NUMERIC}, and 31 cases belong to \textit{DESCRIPTION}. We further analyze the \textit{NUMERIC} question type and find that many parts of the errors are due to higher weights on units such as ``million" or ``years." There exist a total of ten error cases for this type, and we believe that there is room for improvement with regard to these errors through post-processing. 
In the case of the \textit{DESCRIPTION} question type, 17 out of 31 cases are due to inappropriate importance weights. We speculate this result is because the keyphrases for the answers to questions belonging to the \textit{DESCRIPTION} type are sometimes vague; thus, the entire answer needs to be considered when it is evaluated.

\begin{table}[!t]
\small
\centering
\begin{tabular}
{L{0.35\columnwidth}C{0.25\columnwidth}C{0.2\columnwidth}}
\toprule
\textbf{Metrics} & \textbf{MS-MARCO} & \textbf{AVSD} \\ 
\midrule
\textbf{BLEU-1} & 63.44 & 72.02 \\ 
\textbf{ROUGE-L} & 61.29 & 70.98 \\ 
\textbf{BERTScore} & 67.74 & 78.24 \\ 
\midrule
\textbf{BLEU-1-KPQA} & 74.19 & \textbf{81.35} \\ 
\textbf{ROUGE-KPQA} & \textbf{76.34} & 77.20  \\ 
\textbf{BERTScore-KPQA} & \textbf{76.34} & \textbf{81.35} \\ 
\bottomrule 
\end{tabular}
\caption{
The percentage of matches at which human judgment and various metrics on ranking two models' output.
}
\label{table_winlose}
\end{table}

\paragraph{Rank-Pair:}
One practical usage of the text evaluation metric is ranking outputs of multiple models. Using the collected human judgments of correctness for the same 500 \{\textit{question}, \textit{reference answer}\} pairs for two models on MS-MARCO and AVSD, we can compare the output of each models through the human-annotated score. 
To see the alignment of ranking ability among the various metrics with that of human judges, we conduct a ``win-lose match" experiment, counting the number of times that a metric ranks the output of two models as the same as human judges.
To prepare test samples, we chose only those whose gap between human judgment scores on the two models is greater than 2. Finally, we obtain 93 and 193 samples for MS-MARCO and AVSD, respectively. Considering that the range of scores is 1-5, this approach ensures that each output of the models has a clear quality difference.
Table~\ref{table_winlose} shows the percentage of rank-pair matches for each metric with human judgments of correctness on two datasets.
Our KPQA-metric shows more matches than previous metrics in all of the datasets; thus, it is more useful for comparing the generated answers from different models.
\label{sec_ranking}

\section{Related Work}
\label{sec:relatedwork}
One important next step for current QA systems is to generate answers in natural language for a given question and context. Following this interest, several generative (abstractive) QA datasets~\cite{bajaj2016ms, he2018dureader, kovcisky2018narrativeqa, fan2019eli5}, where the answer is not necessarily in the passage, have recently been released.
Since the task is to generate natural language for the given question, the QA system is often trained with seq2seq~\cite{sutskever2014sequence} objective similarly to other natural generation tasks such as neural machine translation. Hence, researchers often use n-gram based similarity metrics such as BLEU to evaluate the GenQA systems, following other natural language generation tasks.
\newline
\indent However, most of these n-gram metrics including BLEU were originally developed to evaluate machine translation and previous works~\cite{liu2016not, nema2018towards, kryscinski2019neural} have shown that these metrics have poor correlations with human judgments in other language generation tasks such as dialogue systems. As with other text generation systems, for GenQA, it is difficult to assess the performance through n-gram metrics. Especially, n-gram similarity metrics can give a high score to a generated answer that is incorrect but shares many unnecessary words with the reference answer. Previous works~\cite{marton2006nuggeteer, yang2018adaptations, chen2019evaluating} have pointed out the difficulty of similar problems and studied automated metrics for evaluating QA systems. Inspired by these works, we focus on studying and developing evaluation metrics for GenQA datasets that have more abstractive and diverse answers. We analyze the problem of using existing n-gram similarity metrics across multiple GenQA datasets and propose alternative metrics for GenQA. 

\section{Conclusion}
\label{sec:conclusion}
In this paper, we create high-quality human judgments on two GenQA datasets, MS-MARCO and AVSD, and show that previous evaluation metrics are poorly correlated with human judgments in terms of the correctness of an answer.
We propose KPQA-metric, which uses the pre-trained model that can predict the importance weights of words in answers to a given question to be integrated with existing metrics.
Our approach has a dramatically higher correlation with human judgments than existing metrics, showing that our model-based importance weighting is critical to measure the correctness of a generated answer in GenQA. 

\section*{Ethical Considerations}
Our paper and dataset follow ethical standards. We compensate the annotators with competitive pay. Furthermore, we follow all ethical procedures for data collection, where we use public datasets to train the models.

\section*{Acknowledgements}
K. Jung is with ASRI, Seoul National University, Korea. This work was supported by AIRS Company in Hyundai  Motor Company \& Kia Corporation through HKMC-SNU AI Consortium Fund.

\bibliography{naacl2021}

\begin{thebibliography}{40}
\expandafter\ifx\csname natexlab\endcsname\relax\def\natexlab#1{#1}\fi

\bibitem[{Alamri et~al.(2019)Alamri, Cartillier, Das, Wang, Cherian, Essa,
  Batra, Marks, Hori, Anderson, Lee, and Parikh}]{alamri2019audio}
Huda Alamri, Vincent Cartillier, Abhishek Das, Jue Wang, Anoop Cherian, Irfan
  Essa, Dhruv Batra, Tim~K. Marks, Chiori Hori, Peter Anderson, Stefan Lee, and
  Devi Parikh. 2019.
\newblock \href {https://doi.org/10.1109/CVPR.2019.00774} {Audio visual
  scene-aware dialog}.
\newblock In \emph{{IEEE} Conference on Computer Vision and Pattern
  Recognition, {CVPR} 2019, Long Beach, CA, USA, June 16-20, 2019}, pages
  7558--7567. Computer Vision Foundation / {IEEE}.

\bibitem[{Alamri et~al.(2018)Alamri, Hori, Marks, Batra, and
  Parikh}]{alamri2018audio}
Huda Alamri, Chiori Hori, Tim~K Marks, Dhruv Batra, and Devi Parikh. 2018.
\newblock Audio visual scene-aware dialog (avsd) track for natural language
  generation in dstc7.
\newblock In \emph{DSTC7 at AAAI2019 Workshop}, volume~2.

\bibitem[{Bajaj et~al.(2016)Bajaj, Campos, Craswell, Deng, Gao, Liu, Majumder,
  McNamara, Mitra, Nguyen et~al.}]{bajaj2016ms}
Payal Bajaj, Daniel Campos, Nick Craswell, Li~Deng, Jianfeng Gao, Xiaodong Liu,
  Rangan Majumder, Andrew McNamara, Bhaskar Mitra, Tri Nguyen, et~al. 2016.
\newblock Ms marco: A human generated machine reading comprehension dataset.
\newblock \emph{arXiv preprint arXiv:1611.09268}.

\bibitem[{Banerjee and Lavie(2005)}]{banerjee2005meteor}
Satanjeev Banerjee and Alon Lavie. 2005.
\newblock \href {https://www.aclweb.org/anthology/W05-0909} {{METEOR}: An
  automatic metric for {MT} evaluation with improved correlation with human
  judgments}.
\newblock In \emph{Proceedings of the {ACL} Workshop on Intrinsic and Extrinsic
  Evaluation Measures for Machine Translation and/or Summarization}, pages
  65--72, Ann Arbor, Michigan. Association for Computational Linguistics.

\bibitem[{Bauer et~al.(2018)Bauer, Wang, and Bansal}]{bauer2018commonsense}
Lisa Bauer, Yicheng Wang, and Mohit Bansal. 2018.
\newblock \href {https://doi.org/10.18653/v1/D18-1454} {Commonsense for
  generative multi-hop question answering tasks}.
\newblock In \emph{Proceedings of the 2018 Conference on Empirical Methods in
  Natural Language Processing}, pages 4220--4230, Brussels, Belgium.
  Association for Computational Linguistics.

\bibitem[{Bi et~al.(2019)Bi, Wu, Yan, Wang, Xia, and Li}]{bi2019incorporating}
Bin Bi, Chen Wu, Ming Yan, Wei Wang, Jiangnan Xia, and Chenliang Li. 2019.
\newblock \href {https://doi.org/10.18653/v1/D19-1255} {Incorporating external
  knowledge into machine reading for generative question answering}.
\newblock In \emph{Proceedings of the 2019 Conference on Empirical Methods in
  Natural Language Processing and the 9th International Joint Conference on
  Natural Language Processing (EMNLP-IJCNLP)}, pages 2521--2530, Hong Kong,
  China. Association for Computational Linguistics.

\bibitem[{Bi et~al.(2020)Bi, Wu, Yan, Wang, Xia, and Li}]{bi2020generating}
Bin Bi, Chen Wu, Ming Yan, Wei Wang, Jiangnan Xia, and Chenliang Li. 2020.
\newblock \href {https://aaai.org/ojs/index.php/AAAI/article/view/6238}
  {Generating well-formed answers by machine reading with stochastic selector
  networks}.
\newblock In \emph{The Thirty-Fourth {AAAI} Conference on Artificial
  Intelligence, {AAAI} 2020, The Thirty-Second Innovative Applications of
  Artificial Intelligence Conference, {IAAI} 2020, The Tenth {AAAI} Symposium
  on Educational Advances in Artificial Intelligence, {EAAI} 2020, New York,
  NY, USA, February 7-12, 2020}, pages 7424--7431. {AAAI} Press.

\bibitem[{Chen et~al.(2019)Chen, Stanovsky, Singh, and
  Gardner}]{chen2019evaluating}
Anthony Chen, Gabriel Stanovsky, Sameer Singh, and Matt Gardner. 2019.
\newblock \href {https://doi.org/10.18653/v1/D19-5817} {Evaluating question
  answering evaluation}.
\newblock In \emph{Proceedings of the 2nd Workshop on Machine Reading for
  Question Answering}, pages 119--124, Hong Kong, China. Association for
  Computational Linguistics.

\bibitem[{Devlin et~al.(2019)Devlin, Chang, Lee, and
  Toutanova}]{devlin2019bert}
Jacob Devlin, Ming-Wei Chang, Kenton Lee, and Kristina Toutanova. 2019.
\newblock \href {https://doi.org/10.18653/v1/N19-1423} {{BERT}: Pre-training of
  deep bidirectional transformers for language understanding}.
\newblock In \emph{Proceedings of the 2019 Conference of the North {A}merican
  Chapter of the Association for Computational Linguistics: Human Language
  Technologies, Volume 1 (Long and Short Papers)}, pages 4171--4186,
  Minneapolis, Minnesota. Association for Computational Linguistics.

\bibitem[{Dong et~al.(2019)Dong, Yang, Wang, Wei, Liu, Wang, Gao, Zhou, and
  Hon}]{dong2019unified}
Li~Dong, Nan Yang, Wenhui Wang, Furu Wei, Xiaodong Liu, Yu~Wang, Jianfeng Gao,
  Ming Zhou, and Hsiao{-}Wuen Hon. 2019.
\newblock \href
  {https://proceedings.neurips.cc/paper/2019/hash/c20bb2d9a50d5ac1f713f8b34d9aac5a-Abstract.html}
  {Unified language model pre-training for natural language understanding and
  generation}.
\newblock In \emph{Advances in Neural Information Processing Systems 32: Annual
  Conference on Neural Information Processing Systems 2019, NeurIPS 2019,
  December 8-14, 2019, Vancouver, BC, Canada}, pages 13042--13054.

\bibitem[{Fan et~al.(2019)Fan, Jernite, Perez, Grangier, Weston, and
  Auli}]{fan2019eli5}
Angela Fan, Yacine Jernite, Ethan Perez, David Grangier, Jason Weston, and
  Michael Auli. 2019.
\newblock \href {https://doi.org/10.18653/v1/P19-1346} {{ELI}5: Long form
  question answering}.
\newblock In \emph{Proceedings of the 57th Annual Meeting of the Association
  for Computational Linguistics}, pages 3558--3567, Florence, Italy.
  Association for Computational Linguistics.

\bibitem[{Gers et~al.(2000)Gers, Schmidhuber, and Cummins}]{gers2000learning}
Felix~A Gers, J{\"u}rgen Schmidhuber, and Fred Cummins. 2000.
\newblock Learning to forget: Continual prediction with lstm.
\newblock \emph{Neural Computation}, 12(10):2451--2471.

\bibitem[{He et~al.(2018)He, Liu, Liu, Lyu, Zhao, Xiao, Liu, Wang, Wu, She,
  Liu, Wu, and Wang}]{he2018dureader}
Wei He, Kai Liu, Jing Liu, Yajuan Lyu, Shiqi Zhao, Xinyan Xiao, Yuan Liu,
  Yizhong Wang, Hua Wu, Qiaoqiao She, Xuan Liu, Tian Wu, and Haifeng Wang.
  2018.
\newblock \href {https://doi.org/10.18653/v1/W18-2605} {{D}u{R}eader: a
  {C}hinese machine reading comprehension dataset from real-world
  applications}.
\newblock In \emph{Proceedings of the Workshop on Machine Reading for Question
  Answering}, pages 37--46, Melbourne, Australia. Association for Computational
  Linguistics.

\bibitem[{Hori et~al.(2017)Hori, Hori, Lee, Zhang, Harsham, Hershey, Marks, and
  Sumi}]{hori2017attention}
Chiori Hori, Takaaki Hori, Teng{-}Yok Lee, Ziming Zhang, Bret Harsham, John~R.
  Hershey, Tim~K. Marks, and Kazuhiko Sumi. 2017.
\newblock \href {https://doi.org/10.1109/ICCV.2017.450} {Attention-based
  multimodal fusion for video description}.
\newblock In \emph{{IEEE} International Conference on Computer Vision, {ICCV}
  2017, Venice, Italy, October 22-29, 2017}, pages 4203--4212. {IEEE} Computer
  Society.

\bibitem[{Joshi et~al.(2020)Joshi, Chen, Liu, Weld, Zettlemoyer, and
  Levy}]{joshi2020spanbert}
Mandar Joshi, Danqi Chen, Yinhan Liu, Daniel~S. Weld, Luke Zettlemoyer, and
  Omer Levy. 2020.
\newblock \href {https://doi.org/10.1162/tacl_a_00300} {{S}pan{BERT}: Improving
  pre-training by representing and predicting spans}.
\newblock \emph{Transactions of the Association for Computational Linguistics},
  8:64--77.

\bibitem[{Jung and Lease(2011)}]{jung2011improving}
Hyun~Joon Jung and Matthew Lease. 2011.
\newblock Improving consensus accuracy via z-score and weighted voting.
\newblock In \emph{Workshops at the Twenty-Fifth AAAI Conference on Artificial
  Intelligence}.

\bibitem[{Ko{\v{c}}isk{\'y} et~al.(2018)Ko{\v{c}}isk{\'y}, Schwarz, Blunsom,
  Dyer, Hermann, Melis, and Grefenstette}]{kovcisky2018narrativeqa}
Tom{\'a}{\v{s}} Ko{\v{c}}isk{\'y}, Jonathan Schwarz, Phil Blunsom, Chris Dyer,
  Karl~Moritz Hermann, G{\'a}bor Melis, and Edward Grefenstette. 2018.
\newblock \href {https://doi.org/10.1162/tacl_a_00023} {The {N}arrative{QA}
  reading comprehension challenge}.
\newblock \emph{Transactions of the Association for Computational Linguistics},
  6:317--328.

\bibitem[{Krippendorff(1970)}]{krippendorff1970estimating}
Klaus Krippendorff. 1970.
\newblock Estimating the reliability, systematic error and random error of
  interval data.
\newblock \emph{Educational and Psychological Measurement}, 30(1):61--70.

\bibitem[{Krippendorff(2011)}]{krippendorff2011computing}
Klaus Krippendorff. 2011.
\newblock Computing krippendorff's alpha-reliability.

\bibitem[{Kryscinski et~al.(2019)Kryscinski, Keskar, McCann, Xiong, and
  Socher}]{kryscinski2019neural}
Wojciech Kryscinski, Nitish~Shirish Keskar, Bryan McCann, Caiming Xiong, and
  Richard Socher. 2019.
\newblock \href {https://doi.org/10.18653/v1/D19-1051} {Neural text
  summarization: A critical evaluation}.
\newblock In \emph{Proceedings of the 2019 Conference on Empirical Methods in
  Natural Language Processing and the 9th International Joint Conference on
  Natural Language Processing (EMNLP-IJCNLP)}, pages 540--551, Hong Kong,
  China. Association for Computational Linguistics.

\bibitem[{Landis and Koch(1977)}]{landis1977measurement}
J~Richard Landis and Gary~G Koch. 1977.
\newblock The measurement of observer agreement for categorical data.
\newblock \emph{biometrics}, pages 159--174.

\bibitem[{Le et~al.(2019)Le, Sahoo, Chen, and Hoi}]{le2019multimodal}
Hung Le, Doyen Sahoo, Nancy Chen, and Steven Hoi. 2019.
\newblock \href {https://doi.org/10.18653/v1/P19-1564} {Multimodal transformer
  networks for end-to-end video-grounded dialogue systems}.
\newblock In \emph{Proceedings of the 57th Annual Meeting of the Association
  for Computational Linguistics}, pages 5612--5623, Florence, Italy.
  Association for Computational Linguistics.

\bibitem[{Likert(1932)}]{likert1932technique}
Rensis Likert. 1932.
\newblock A technique for the measurement of attitudes.
\newblock \emph{Archives of psychology}.

\bibitem[{Lin(2004)}]{lin-2004-rouge}
Chin-Yew Lin. 2004.
\newblock \href {https://www.aclweb.org/anthology/W04-1013} {{ROUGE}: A package
  for automatic evaluation of summaries}.
\newblock In \emph{Text Summarization Branches Out}, pages 74--81, Barcelona,
  Spain. Association for Computational Linguistics.

\bibitem[{Liu et~al.(2016)Liu, Lowe, Serban, Noseworthy, Charlin, and
  Pineau}]{liu2016not}
Chia-Wei Liu, Ryan Lowe, Iulian Serban, Mike Noseworthy, Laurent Charlin, and
  Joelle Pineau. 2016.
\newblock \href {https://doi.org/10.18653/v1/D16-1230} {How {NOT} to evaluate
  your dialogue system: An empirical study of unsupervised evaluation metrics
  for dialogue response generation}.
\newblock In \emph{Proceedings of the 2016 Conference on Empirical Methods in
  Natural Language Processing}, pages 2122--2132, Austin, Texas. Association
  for Computational Linguistics.

\bibitem[{Loshchilov and Hutter(2018)}]{loshchilov2018decoupled}
Ilya Loshchilov and Frank Hutter. 2018.
\newblock Decoupled weight decay regularization.
\newblock In \emph{International Conference on Learning Representations}.

\bibitem[{Marton and Radul(2006)}]{marton2006nuggeteer}
Gregory Marton and Alexey Radul. 2006.
\newblock \href {https://www.aclweb.org/anthology/N06-1048} {{N}uggeteer:
  Automatic nugget-based evaluation using descriptions and judgements}.
\newblock In \emph{Proceedings of the Human Language Technology Conference of
  the {NAACL}, Main Conference}, pages 375--382, New York City, USA.
  Association for Computational Linguistics.

\bibitem[{Nema and Khapra(2018)}]{nema2018towards}
Preksha Nema and Mitesh~M. Khapra. 2018.
\newblock \href {https://doi.org/10.18653/v1/D18-1429} {Towards a better metric
  for evaluating question generation systems}.
\newblock In \emph{Proceedings of the 2018 Conference on Empirical Methods in
  Natural Language Processing}, pages 3950--3959, Brussels, Belgium.
  Association for Computational Linguistics.

\bibitem[{Nishida et~al.(2019)Nishida, Saito, Nishida, Shinoda, Otsuka, Asano,
  and Tomita}]{nishida2019multi}
Kyosuke Nishida, Itsumi Saito, Kosuke Nishida, Kazutoshi Shinoda, Atsushi
  Otsuka, Hisako Asano, and Junji Tomita. 2019.
\newblock \href {https://doi.org/10.18653/v1/P19-1220} {Multi-style generative
  reading comprehension}.
\newblock In \emph{Proceedings of the 57th Annual Meeting of the Association
  for Computational Linguistics}, pages 2273--2284, Florence, Italy.
  Association for Computational Linguistics.

\bibitem[{Ostermann et~al.(2018)Ostermann, Roth, Modi, Thater, and
  Pinkal}]{ostermann2018semeval}
Simon Ostermann, Michael Roth, Ashutosh Modi, Stefan Thater, and Manfred
  Pinkal. 2018.
\newblock \href {https://doi.org/10.18653/v1/S18-1119} {{S}em{E}val-2018 task
  11: Machine comprehension using commonsense knowledge}.
\newblock In \emph{Proceedings of The 12th International Workshop on Semantic
  Evaluation}, pages 747--757, New Orleans, Louisiana. Association for
  Computational Linguistics.

\bibitem[{Papineni et~al.(2002)Papineni, Roukos, Ward, and
  Zhu}]{papineni-etal-2002-bleu}
Kishore Papineni, Salim Roukos, Todd Ward, and Wei-Jing Zhu. 2002.
\newblock \href {https://doi.org/10.3115/1073083.1073135} {{B}leu: a method for
  automatic evaluation of machine translation}.
\newblock In \emph{Proceedings of the 40th Annual Meeting of the Association
  for Computational Linguistics}, pages 311--318, Philadelphia, Pennsylvania,
  USA. Association for Computational Linguistics.

\bibitem[{Rajpurkar et~al.(2016)Rajpurkar, Zhang, Lopyrev, and
  Liang}]{rajpurkar2016squad}
Pranav Rajpurkar, Jian Zhang, Konstantin Lopyrev, and Percy Liang. 2016.
\newblock \href {https://doi.org/10.18653/v1/D16-1264} {{SQ}u{AD}: 100,000+
  questions for machine comprehension of text}.
\newblock In \emph{Proceedings of the 2016 Conference on Empirical Methods in
  Natural Language Processing}, pages 2383--2392, Austin, Texas. Association
  for Computational Linguistics.

\bibitem[{Song et~al.(2017)Song, Wang, and Hamza}]{song2017unified}
Linfeng Song, Zhiguo Wang, and Wael Hamza. 2017.
\newblock A unified query-based generative model for question generation and
  question answering.
\newblock \emph{arXiv preprint arXiv:1709.01058}.

\bibitem[{Sutskever et~al.(2014)Sutskever, Vinyals, and
  Le}]{sutskever2014sequence}
Ilya Sutskever, Oriol Vinyals, and Quoc~V. Le. 2014.
\newblock \href
  {https://proceedings.neurips.cc/paper/2014/hash/a14ac55a4f27472c5d894ec1c3c743d2-Abstract.html}
  {Sequence to sequence learning with neural networks}.
\newblock In \emph{Advances in Neural Information Processing Systems 27: Annual
  Conference on Neural Information Processing Systems 2014, December 8-13 2014,
  Montreal, Quebec, Canada}, pages 3104--3112.

\bibitem[{Vedantam et~al.(2015)Vedantam, Zitnick, and
  Parikh}]{vedantam2015cider}
Ramakrishna Vedantam, C.~Lawrence Zitnick, and Devi Parikh. 2015.
\newblock \href {https://doi.org/10.1109/CVPR.2015.7299087} {Cider:
  Consensus-based image description evaluation}.
\newblock In \emph{{IEEE} Conference on Computer Vision and Pattern
  Recognition, {CVPR} 2015, Boston, MA, USA, June 7-12, 2015}, pages
  4566--4575. {IEEE} Computer Society.

\bibitem[{Wolf et~al.(2019)Wolf, Debut, Sanh, Chaumond, Delangue, Moi, Cistac,
  Rault, Louf, Funtowicz, and Brew}]{Wolf2019HuggingFacesTS}
Thomas Wolf, Lysandre Debut, Victor Sanh, Julien Chaumond, Clement Delangue,
  Anthony Moi, Pierric Cistac, Tim Rault, R'emi Louf, Morgan Funtowicz, and
  Jamie Brew. 2019.
\newblock Huggingface's transformers: State-of-the-art natural language
  processing.
\newblock \emph{ArXiv}, abs/1910.03771.

\bibitem[{Yang et~al.(2018{\natexlab{a}})Yang, Liu, Liu, Lyu, and
  Li}]{yang2018adaptations}
An~Yang, Kai Liu, Jing Liu, Yajuan Lyu, and Sujian Li. 2018{\natexlab{a}}.
\newblock \href {https://doi.org/10.18653/v1/W18-2611} {Adaptations of {ROUGE}
  and {BLEU} to better evaluate machine reading comprehension task}.
\newblock In \emph{Proceedings of the Workshop on Machine Reading for Question
  Answering}, pages 98--104, Melbourne, Australia. Association for
  Computational Linguistics.

\bibitem[{Yang et~al.(2018{\natexlab{b}})Yang, Qi, Zhang, Bengio, Cohen,
  Salakhutdinov, and Manning}]{yang2018hotpotqa}
Zhilin Yang, Peng Qi, Saizheng Zhang, Yoshua Bengio, William Cohen, Ruslan
  Salakhutdinov, and Christopher~D Manning. 2018{\natexlab{b}}.
\newblock Hotpotqa: A dataset for diverse, explainable multi-hop question
  answering.
\newblock In \emph{Proceedings of the 2018 Conference on Empirical Methods in
  Natural Language Processing}, pages 2369--2380.

\bibitem[{Yin et~al.(2016)Yin, Jiang, Lu, Shang, Li, and Li}]{yin2016neural}
Jun Yin, Xin Jiang, Zhengdong Lu, Lifeng Shang, Hang Li, and Xiaoming Li. 2016.
\newblock \href {http://www.ijcai.org/Abstract/16/422} {Neural generative
  question answering}.
\newblock In \emph{Proceedings of the Twenty-Fifth International Joint
  Conference on Artificial Intelligence, {IJCAI} 2016, New York, NY, USA, 9-15
  July 2016}, pages 2972--2978. {IJCAI/AAAI} Press.

\bibitem[{Zhang et~al.(2020)Zhang, Kishore, Wu, Weinberger, and
  Artzi}]{zhang2020bertscore}
Tianyi Zhang, Varsha Kishore, Felix Wu, Kilian~Q. Weinberger, and Yoav Artzi.
  2020.
\newblock \href {https://openreview.net/forum?id=SkeHuCVFDr} {Bertscore:
  Evaluating text generation with {BERT}}.
\newblock In \emph{8th International Conference on Learning Representations,
  {ICLR} 2020, Addis Ababa, Ethiopia, April 26-30, 2020}. OpenReview.net.

\end{thebibliography}
\bibliographystyle{acl_natbib}

\appendix



\begin{figure*}[t]
\small
\centering
\includegraphics[width=2\columnwidth]{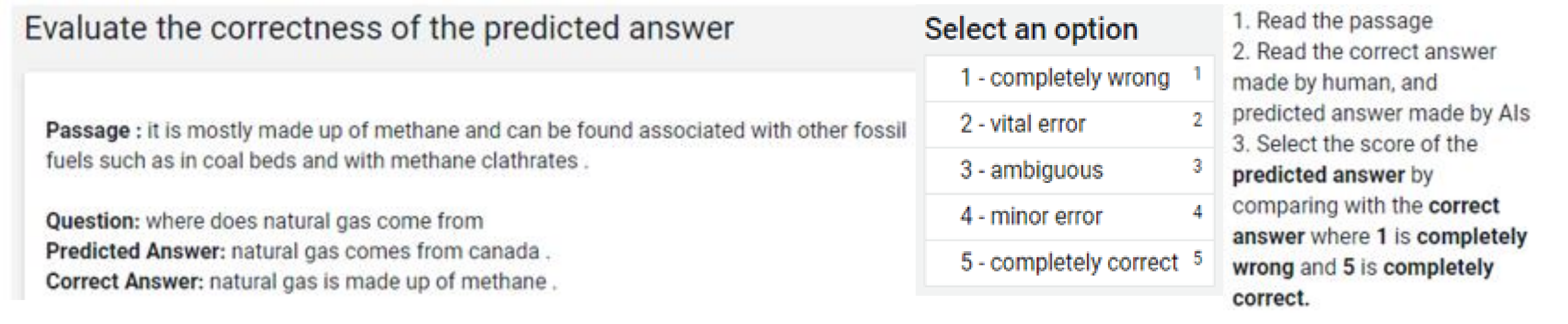}
\caption{
 Instruction for MTurk workers}
\label{fig_guide}
\end{figure*}

\section{Data Collection}
\label{appendix:collect}

\subsection{Datasets}
\label{appendix:dataset}
We collect human judgments of correctness for two GenQA datasets, MS-MARCO~\cite{bajaj2016ms} and AVSD~\cite{alamri2019audio}. We describe the properties of each dataset in this section.

\paragraph{MS-MARCO} MS-MARCO is a large-scale english machine reading comprehension dataset that provides ten candidate passages for each question. The model should consider the relevance of the passages for the given question and answer the question. One of the main features of this dataset is that it contains free-form answers that are abstractive. MS-MARCO provides two tasks, Natural Language Generation (NLG) task and Q\&A task. For the NLG task, the model should generate an abstractive summary of the passages for given questions, which is a well-formed answer rather than an answer span in the passage. Although the Q\&A task also provides some abstractive answers, most of the answers are short and do not contain the context or rationale of the question. 
Hence, we use the NLG subset of MS-MARCO dataset as a GenQA dataset to study the metrics for GenQA. Also, we use the training set of Q\&A subset to train and evaluate KPQA, since most of the samples in this subset has exact answer spans in the passage like SQuAD.

\paragraph{Audio Visual Scene-aware Dialog (AVSD)} To study more general metrics for GenQA, we also use a multimodal GenQA dataset for our work. Audio Visual Scene-aware Dialog (AVSD) is a multimodal dialogue dataset composed of QA pair about Charades videos. Although the name of the dataset contains dialog, all of the dialog pairs are composed of questions answering about a video. The task of this dataset is to generate an answer for a question about a given video, audio, and the history of previous turns in the dialog. In other words, this task is to generate a free-form answer for a given multimodal context, which can be considered as GenQA.

\begin{table}[t]
\small
\centering
\begin{tabular}
{L{0.23\columnwidth}C{0.20\columnwidth}C{0.17\columnwidth}C{0.2\columnwidth}}
\toprule
\textbf{Dataset} & \textbf{Model} & \textbf{BLEU-1} & \textbf{ROUGE-L}\\
\midrule
\multirow{2}{*}{MS-MARCO} & UniLM & 60.2 & 63.1\\
                          & MHPGM & 43.7 & 53.9\\
\midrule
\multirow{2}{*}{AVSD} & MTN & 67.3 & 52.6\\
                      & AMF & 62.6 & 48.7\\        
\bottomrule 
\end{tabular}
\caption{
Performance of the model we trained to generate answers
}
\label{table_gen_ans}
\end{table}
\subsection{Instructions to Annotators}

The full instructions to annotators in MTurk are shown in Figure~\ref{fig_guide}. We hire the annotators whose HIT approval rate are higher than 95\% and pay \$0.03 for each assignment.  

\subsection{Models}
\label{appendix:model}
To investigate the performance of automatic metrics, we gather pairs of a sentence, \{generated answer, \textit{reference answer}\}.
Collecting high-quality answer candidates for a given context and question is an essential step; thus, we choose two models for each dataset from the latest research in the literature. We train two models UniLM~\cite{dong2019unified} and MHPGM~\cite{bauer2018commonsense} for MS-MARCO dataset. For AVSD dataset, we train two models MTN~\cite{le2019multimodal} and AMF~\cite{alamri2018audio}. We present the performance of each model we trained in Table~\ref{table_gen_ans}. 
We briefly describe the models and the training details to generate the answer for two datasets.

\paragraph{UniLM} UniLM, which stands for unified language model pre-training, is a powerful seq2seq model based on pre-trained representations from BERT~\cite{devlin2019bert}. UniLM is a pre-trained transformer network that can be easily fine-tuned for NLU and NLG. UniLM achieves higher performance for various NLG tasks, such as abstractive summarization and question generation. We fine-tune UniLM for GenQA similar to the way fine-tuning UniLM to NLG, where
source sequences are each question and paragraphs, the target sequence is an answer. We add [SEP] tokens between the question and each paragraph. Then, we fine-tune UniLM for 3 epochs with this setting using the public code\footnote{https://github.com/microsoft/unilm}.

\paragraph{MHPGM} MHPGM, which stands for multi-hop pointer generator networks, uses multi-hop reasoning QA model that can integrate commonsense information. This model uses pointer-generator decoder to generate the answer. We train the model for three epochs with batch size 24 using the public code\footnote{https://github.com/yicheng-w/CommonSenseMultiHopQA}. 

\begin{table*}[t]
\small
\centering
\begin{tabular}
{L{0.35\columnwidth}C{0.15\columnwidth}C{0.15\columnwidth}C{0.15\columnwidth}C{0.15\columnwidth}
C{0.15\columnwidth}C{0.15\columnwidth}C{0.15\columnwidth}C{0.15\columnwidth}}
\toprule

\multicolumn{1}{c}{\textbf{Dataset}} & \multicolumn{4}{c}{\textbf{MS-MARCO}} 
& \multicolumn{4}{c}{\textbf{AVSD}}\\ 

\midrule

\multicolumn{1}{c}{\textbf{Model}} & \multicolumn{2}{c}{\textbf{UniLM}} & 
\multicolumn{2}{c}{\textbf{MHPGM}} & \multicolumn{2}{c}{\textbf{MTN}} &
\multicolumn{2}{c}{\textbf{AMF}}\\

\cmidrule{1-9}
 \textbf{Metric}
              & \textbf{$r$} & \textbf{$\rho$} 
              & \textbf{$r$} & \textbf{$\rho$}
              & \textbf{$r$} & \textbf{$\rho$}
              & \textbf{$r$} & \textbf{$\rho$}\\
\midrule
\textbf{BLEU-1} & 0.369 & 0.337 & 0.331 & 0.312 & 0.497 & 0.516 & 0.655 & 0.580 \\
\textbf{BLEU-4} & 0.173 & 0.224 & 0.227 & 0.26 & 0.441 & 0.492 & 0.579 & 0.553 \\
\textbf{ROUGE-L} & 0.317 & 0.289 & 0.305 & 0.307 & 0.510 & 0.528 & 0.648 & 0.575 \\
\textbf{METEOR} & 0.431 & 0.408 & 0.425 & 0.422 & 0.521 & 0.596 & 0.633 & 0.608 \\
\textbf{CIDEr} & 0.261 & 0.256 & 0.292 & 0.289 & 0.509 & 0.559 & 0.627 & 0.602 \\
\textbf{BERTScore} & 0.469 & 0.445 & 0.466 & 0.472 & 0.592 & 0.615 & 0.701 & 0.645 \\
\midrule
\textbf{BLEU-1-KPQA} & 0.729 & \textbf{0.678} & 0.612 & 0.573 & 0.687 & 0.681 & 0.736 & 0.673 \\
\textbf{ROUGE-L-KPQA} & \textbf{0.732} & 0.667 & \textbf{0.667} & 0.624 & 0.681 & 0.682 & 0.731 & \textbf{0.700} \\
\textbf{BERTScore-KPQA} & 0.696 & 0.659 & 0.659 & \textbf{0.655} & \textbf{0.712} & \textbf{0.703} & \textbf{0.738} & 0.695 \\

\bottomrule 
\end{tabular}
\caption{
Pearson Correlation($r$) and Spearman's Correlation($\rho$) between various automatic metrics and human judgments of correctness for MS-MARCO dataset and AVSD dataset. We generate the answers and collect human judgments for two models on each dataset. All of the results are statistically significant (p-value $<$ 0.01).
}

\label{comparison-model-type}
\end{table*}
\paragraph{MTN} MTN~\cite{le2019multimodal}, which is a multimodal transformer encoder-decoder framework, is a state-of-the-art model for AVSD. MTN employs multimodal attention blocks to fuse multiple modalities such as text, video, and audio. We train 10 epochs with batch size 256 and generate the answers for the testset released in the DSTC7 workshop~\cite{alamri2018audio} using the publicaly available code\footnote{https://github.com/henryhungle/MTN}.

\paragraph{AMF} AMF is an Attentional Multimodal Fusion based model~\cite{hori2017attention} introduced as a baseline system for DSTC7 AVSD workshop~\cite{alamri2018audio}, It is composed of RNN and multimodal attention architecture. This model encode the multimodal inputs with LSTM~\cite{gers2000learning} and fuse the information with modality-dependent attention mechanism. We train this model with 15 epochs with batch size 64 using the public code\footnote{https://github.com/dialogtekgeek/\\
AudioVisualSceneAwareDialog}.

\section{Further Experiments}
\subsection{Correlation by Models}
The dataset we collect has human judgments on a generated answer from two models for each dataset; thus we can observe how the performance of each metric depends on the type of GenQA model. The experimental results in Table~\ref{comparison-model-type} show that our proposed metric outperforms other metrics in both of the GenQA models for each dataset.

\section{Experimental Details}

In this section, we describe experimental details that are not mentioned in the previous sections including some items in the reproducibility checklist.

\subsection{Reproducibility Checklist}

\paragraph{Source Code}
We provide the source code for both training KPQA and computing KPQA metric as a supplementary material. We will publicly release the full source with the pre-trained model to easily compute KPQA-metric. 

\paragraph{Computing Infrastructure}
We use Intel(R) Core(TM) i7-6850K CPU (3.60 GHz) with GeForce GTX 1080 Ti for the experiments. The software environments are Python 3.6 and PyTorch 1.3.1.

\paragraph{Average runtime for each approach}
Each epoch of our training KPQA on average takes 150 minutes using the single GPU. For evaluation, it takes 5 minutes.

\paragraph{Number of Model Parameters} The number of parameters in KPQA model is about 109.4M.

\paragraph{Hyperparameters}
We use max sequence length of 256 for the inputs of KPQA. We use AdamW~\cite{loshchilov2018decoupled} optimizer with learning rate 2e-5, and mini-batch size of 16 for all of the experiments. We use \textit{bert-base-uncased} with additional one fully-connected layer of 768 units and tanh activation function. And then we add a softmax layer after it. We train KPQA for 5 epochs and choose the model that shows the minimum evaluation loss over the development set. We repeat training 5 times for each best-performing model.

\subsection{Significant Test}
For all of the correlation coefficients we computed in the paper, we use a t-test using a null hypothesis that is an absence of association to report p-value, which is the standard way to test the correlation coefficient.

\begin{table}[h]
\small
\centering
\begin{tabular}
{L{0.4\columnwidth}C{0.15\columnwidth}}
\toprule
\textbf{Dataset} & \textbf{F1}\\ 
\midrule
SQuAD & 55.81 \\
MS-MARCO Q\&A & 59.26 \\
\bottomrule 
\end{tabular}
\caption{
Performance of our keyphrase predictor in development set of each dataset.
}
\label{kp_acc}
\end{table}
\subsection{KPQA Performance}
We present the performance of KPQA on keyphrase prediction for evaluation data in Table~\ref{kp_acc}.

\subsection{BERTScore}
For computing BERTScore we use \textit{bert-large-uncased-whole-word-masking-finetuned-squad} variant from~\cite{Wolf2019HuggingFacesTS}\footnote{https://github.com/huggingface/transformers} which is a BERT model fine-tuned on QA dataset SQuAD. We observe that computing BERTScore through this BERT model shows slightly higher correlation with human judgments than the BERT model without fine tuning. We use the first layer of it after the word embedding layer to compute the embedding. We experiment among different layers and found that the first hidden layer yielded the best result. We compute all of the BERTScore including original BERTScore and BERTScore variants using this BERT model.
\label{sec:appendix}

\end{document}